\documentclass[sigconf]{acmart}
\AtBeginDocument{%
  }

\setcopyright{acmlicensed}
\copyrightyear{2026}
\acmYear{2026}
\setcopyright{cc}
\setcctype{by-nc-nd}
\acmConference[MM '26]{Proceedings of the 34th ACM International Conference on Multimedia}{November 10--14, 2026}{Rio de Janeiro, Brazil}
\acmBooktitle{Proceedings of the 34th ACM International Conference on Multimedia (MM '26), November 10--14, 2026, Rio de Janeiro, Brazil}
\acmDOI{10.1145/3767308.3838639}
\acmISBN{979-8-4007-2213-4/2026/11}

\usepackage{xcolor}
\usepackage{multirow}
\usepackage{subcaption}
\usepackage{balance}

\begin{document}

\title{DDL: A Large-Scale Dataset for Deepfake Detection and Localization in Diversified Real-World Scenarios}

\author{Changtao Miao}
\author{Yi Zhang}
\author{Weize Gao}
\affiliation{%
  \institution{Ant Digital Technologies, Ant Group}
  \city{Hangzhou}
  \country{China}
}

\author{Zhiya Tan}
\affiliation{%
  \institution{Nanyang Technological University}
  \city{Singapore}
  \country{Singapore}
}

\author{Weiwei Feng}
\authornote{Corresponding authors. Email: \{fengweiwei.fww; jianshu.l\}@antgroup.com}
\affiliation{%
  \institution{Ant Group}
  \city{Hangzhou}
  \country{China}
}

\author{Jianshu Li}
\authornotemark[1]
\author{Man Luo}
\affiliation{%
  \institution{Ant Digital Technologies, Ant Group}
  \city{Hangzhou}
  \country{China}
}

\author{Ajian Liu}
\affiliation{%
  \institution{Institute of Automation, Chinese Academy of Sciences}
  \city{Beijing}
  \country{China}
}

\author{Yunfeng Diao}
\affiliation{%
  \institution{Hefei University of Technology}
  \city{Hefei}
  \country{China}
}

\author{Qi Chu}
\author{Tao Gong}
\affiliation{%
  \institution{Anhui Province Key Laboratory of Digital Security}
  \city{Hefei}
  \country{China}
}

\author{Zhe Li}
\author{Weibin Yao}
\affiliation{%
  \institution{Ant Digital Technologies, Ant Group}
  \city{Hangzhou}
  \country{China}
}

\author{Joey Tianyi Zhou}
\affiliation{%
  \institution{Singapore Management University}
  \institution{IAIC, Agency for Science, Technology and Research}
  \city{Singapore}
  \country{Singapore}
}

\renewcommand{\shortauthors}{Changtao Miao et al.}

\begin{abstract}
Recent advances in AIGC have exacerbated the misuse of malicious deepfake content, making the development of reliable deepfake detection methods an essential means to address this challenge. 
Although existing deepfake detection models demonstrate outstanding performance in detection metrics, most methods only provide simple binary classification results, lacking interpretability. 
Recent studies have attempted to enhance the interpretability of classification results by providing spatial manipulation masks or temporal forgery segments. However, due to the limitations of forgery datasets, the practical effectiveness of these methods remains suboptimal. 
The primary reason lies in the fact that most existing deepfake datasets contain only binary labels, with limited variety in forgery scenarios, insufficient diversity in deepfake types, and relatively small data scales, making them inadequate for complex real-world scenarios.
To address this predicament, we construct a novel large-scale deepfake detection and localization (\textbf{DDL}) dataset containing $\textbf{1.4M+}$ forged samples and encompassing $\textbf{80}$ distinct deepfake methods. 
The DDL design incorporates four key innovations: (1) \textbf{Comprehensive Deepfake Methods} (covering 7 different generation architectures and a total of 80 methods), (2) \textbf{Varied Manipulation Modes} (incorporating 7 classic and 3 novel forgery modes), (3) \textbf{Diverse Forgery Scenarios and Modalities}  (including 3 scenarios and 3 modalities),  and (4) \textbf{Fine-grained Forgery Annotations} (providing 1.18M+ precise spatial masks and 0.23M+ precise temporal segments).
Through these improvements, our DDL not only provides a more challenging benchmark for complex real-world forgeries but also offers crucial support for building next-generation deepfake detection, localization, and interpretability methods.
The dataset and supplemental materials are publicly available at {https://github.com/inclusionConf/DDL}.
\end{abstract}


\maketitle

\begin{figure*}[tp]
    \centering
  \includegraphics[width=\textwidth]{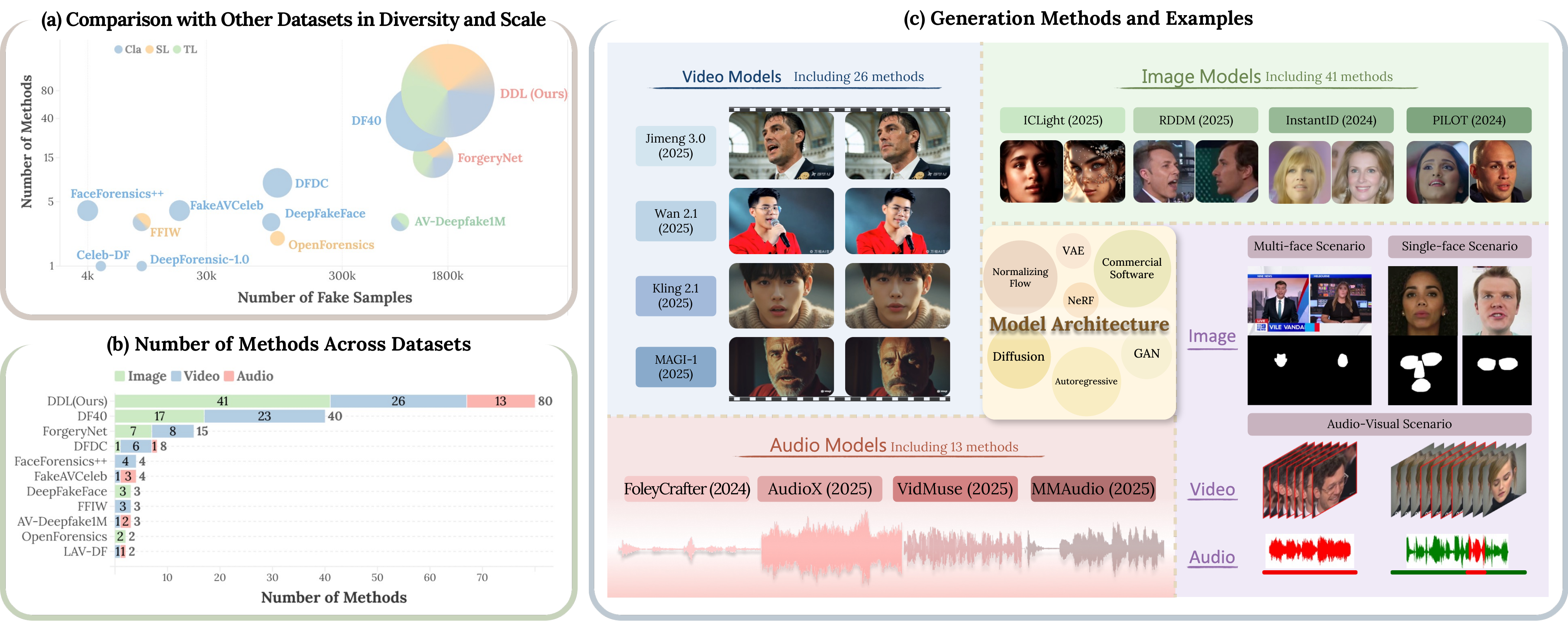}
  \caption{Overview of our DDL dataset. It shows the DDL's advantages in comprehensive deepfake methods, varied manipulation modes, diverse forgery scenarios and modalities, and fine-grained forgery annotations.}
  \Description{}
  \label{fig:datasets1}
\end{figure*}

\section{Introduction}
\label{sec:intro}

The rapid advancement of Artificial Intelligence Generated Content (AIGC) technologies have demonstrated exceptional capabilities in visual content synthesis and editing, finding widespread application in the film and entertainment industries. However, these advancements coexist with severe misuse risks, particularly in generating deepfake images/videos with malicious intent, such as manipulated faces and fabricated news dissemination. 
Consequently, developing a comprehensive and reliable evaluation benchmark for deepfake detection methods is critical and urgent.

Previous deepfake detection research efforts explore various innovative approaches \cite{yan2023ucf,oorloff2024avff,lin2024fake,sun2024diffusionfake,diao2026vulnerabilities,miao2021learning,miao2023f,miao2022hierarchical} to improve binary classification capabilities, because most existing datasets just providing 0-1 label for identifying whether the image is fake or real.
However, binary classification results often lack intuitive and convincing justification, primarily because most deepfake detection methods function as black-box models with opaque decision-making processes. 
Therefore, the existing evaluation protocol can confuse users about why an image is deemed fake and which region is manipulated, forcing them to rely on their knowledge to reassess suspicious images, which lacks interpretability of evaluation.
To a certain extent, recent studies employ visualization techniques to highlight potential manipulation regions in forged images for eliminating this confusion. 
While, these qualitative results lack rigorous quantitative analysis capabilities.
In an effort to further improve interpretability, recent methods \cite{kong2022detect,huang2022fakelocator,miao2025multi,zhang2024comics,miao2025mixture,li2026prototype,che2026gifguard,wang2026sapl,chen2026thinking} attempt to generate spatial manipulation masks or temporal forgery segments as supplements to binary classification results. 
Although these methods provide extra predictions, they still struggle to fully address the issue of interpretability.

The core reason lies on that current deepfake datasets predominantly provide image-level or video-level binary classification labels, lacking fine-grained annotations of manipulation.
For instance, the latest DF40 \cite{yandf40} dataset, despite encompassing 40 forgery types and containing 1.1M+ fake samples, only offers image-level binary labels, which are inadequate for forgery localization tasks. 
Some researchers \cite{kong2022detect,miao2025multi} manually annotate the existing FF++ dataset to obtain relatively precise spatial manipulation masks, but this post-processing approach cannot substitute for accurate manipulation region annotations generated based on the actual forgery process.
Further, certain datasets preserve manipulation region mask annotations during the forgery generation process.
For example, the Dolos \cite{tantaru2024dolos} dataset annotates local manipulation region masks in single-face forgery scenarios, while OpenForensics \cite{le2021openforensics} and FFIW \cite{zhou2021face} extend this approach to multi-face scenarios, providing more complex manipulation region annotations. 
However, these datasets are limited in forgery types and do not cover audio-video forgeries. LAV-DF \cite{cai2022you} is the first audio-video temporal manipulation localization dataset, but its scale is relatively small, with only $10K$ samples. 
The AV-Deepfake1M \cite{cai2024av} dataset further expands the scale to millions of samples, but it only includes a single video forgery method and two audio forgery types, severely lacking diversity in fake samples and making it difficult to adapt to complex real-world scenarios. 
Consequently, the current limitations of forgery localization datasets may impose significant constraints on the interpretability capabilities of deepfake detection models.

\begin{table*}[ht]
\caption{Comparison of existing deepfake datasets. Our DDL surpasses others in terms of the diversity of deepfake methods and the scale of fake samples. Cla: Binary classification. SL: Spatial forgery localization. TL: Temporal forgery localization. 
A: Audio modality. I: Image modality. V: Video modality.
SF: Single-Face scenario. MF: Multi-Face scenario. AV: Audio-Visual scenario.
Methods: The total number of deepfake methods in the dataset.
\#Fake: Number of forged samples, each deepfake image, video, or audio is considered as one sample.}
\centering
\scalebox{0.98}{
\begin{tabular}{l|c|c|c|c|c|c|c}
\toprule
Datasets & Publication & Tasks & Modality & Scenarios & Latest Deepfake & Methods & \#Fake \\ 
\midrule
FF++ \cite{rossler2019faceforensics++} & ICCV' 19 & Cla & V & SF & NeuralTextures (2019) & 4 & 4K \\
Celeb-DF \cite{li2020celeb} & CVPR' 20 & Cla & V & SF & Unknown & 1 & 5K+ \\
DF-1.0 \cite{jiang2021deeperforensics} & CVPR' 20 & Cla & V & SF & DF-VAE (2020) & 1 & 10K \\
DFDC \cite{dolhansky2020deepfake} & Arxiv' 20 & Cla & V & SF & StyleGAN (2018) & 8 & 0.1M+ \\
FFIW \cite{zhou2021face} & CVPR' 21 & Cla/SL & V & MF & FSGAN (2019) & 3 & 10K \\
OpenForensics \cite{le2021openforensics} & ICCV' 21 & SL & I & MF & InterFaceGAN (2020) & 2 & 0.1M \\
FakeAVCeleb \cite{khalid2021fakeavceleb} & NeurIPS' 21 & Cla & A/V & SF & Wav2Lip (2021) & 4 & 19K+ \\
ForgeryNet \cite{he2021forgerynet} & CVPR' 21 & Cla/TL/SL & I/V & SF & StarGANv2 (2020) & 15 & 1.4M+ \\
LAV-DF \cite{cai2022you} & DICTA' 22 & Cla/TL & A/V & SF/AV & Wav2Lip (2021) & 2 & 10K \\
DeepFakeFace \cite{song2023robustness} & ArXiv’23 & Cla & I & SF & Stable-Diffusion (2021) & 3 & 90K \\
DiffusionDeepfake \cite{bhattacharyya2024diffusion} & ArXiv’24 & Cla & I & SF & Stable-Diffusion(2021) & 3 & 0.1M+ \\
AV-Deepfake1M \cite{cai2024av} & MM' 24 & Cla/TL & A/V & SF/AV & TalkLip (2023) & 3 & 0.8M+ \\
DF40 \cite{yandf40} & NeurIPS' 24 & Cla & I/V & SF & PixArt-$\alpha$ (2024) & 40 & 1.1M+ \\
MFFI \cite{miao2025mffi} & MM' 25 & Cla & I & SF & Jimeng 3.0 (2025) & 50 & 0.7M+ \\
ArEnAV \cite{kuckreja2025tell} & ICLR' 26 & Cla/TL & A/V & SF & LatentSync (2025) & 6 & 0.3M \\
\midrule
\textbf{DDL (ours)} & \textbf{--} & \textbf{Cla/TL/SL} & \textbf{A/I/V} & \textbf{SF/MF/AV} & \textbf{Kling 2.1 (2025)} & \textbf{80} & \textbf{1.4M+} \\ 
\bottomrule
\end{tabular}
}
\label{tab:datasets_1}
\end{table*}

To address these limitations, we propose a large-scale, diverse and multi-modal Deepfake Detection and Localization (\textbf{DDL}) dataset,   as illustrated in Fig. \ref{fig:datasets1}. 
This dataset encompasses both unimodal images (DDL-I) and multi-modal audio-visual (DDL-AV) content, specifically designed for spatial forgery localization and temporal forgery localization tasks, respectively. 
With a total of $\textbf{1.4M+}$ forged samples, DDL incorporates $\textbf{80}$ state-of-the-art Deepfake techniques and a broader spectrum of technology types, including audio, video, and image forgery methods, resulting in a significantly more diverse and challenging benchmark.
Specifically, we constructed this dataset with the following key innovations: 
\textbf{(1) Comprehensive Deepfake Methods}: The dataset includes 80 Deepfake techniques spanning from common GANs \cite{goodfellow2014generative} and Diffusion models \cite{ho2020denoising} to emerging architectures such as VAEs \cite{kingma2013auto}, Normalizing Flows \cite{kobyzev2020normalizing}, NeRFs \cite{mildenhall2021nerf}, Autoregressive \cite{tian2024visual} models, and popular commercial software. Furthermore, it encompasses both visual and audio modalities, enriching the dataset's technological diversity. 
\textbf{(2) Varied Manipulation Modes}: In the spatial domain, DDL covers face swapping, face reenactment, full-face synthesis, and face editing. In the temporal domain, it includes deletion, replacement, and insertion operations of forged content. Notably, we introduce hybrid face forgery, audio-visual asynchronous manipulation, and audio-visual full synthesis modes for the first time, further increasing complexity and realism. 
\textbf{(3) Diverse Forgery Scenarios and Modalities}: DDL covers single-face, multi-face, and audio-visual scenarios, and incorporates audio, image, and video modalities, simulating complex real-world forgery content.
\textbf{(4) Fine-grained Forgery Annotations}: We provide spatial forgery region masks and temporal forgery segment labels, including precise 1.18M+ spatial masks and 0.23M+
temporal segments. 
These detailed annotations significantly enhance the research capabilities for forgery localization tasks. 
DDL thereby establishes a challenging and practical benchmark for deepfake detection, localization, and interpretation in complex real-world scenarios.

A detailed comparison with existing deepfake datasets is shown in Table \ref{tab:datasets_1}. To our best knowledge, DDL is the first dataset to uniformly consider three modalities (Audio, Image, Video) and three scenarios (Single-Face, Multi-Face, Audio-Visual) for multi-task deepfake detection and localization. By encompassing $80$ diverse techniques and $1.4M+$ forged samples, DDL serves as the core benchmark for the \textit{CodaBench Deepfake Detection and Localization Challenge\footnotemark[1]}, driving robust real-world deepfake detection and localization research. 
\footnotetext[1]{https://www.codabench.org/competitions/6891/}

Besides, it is noted that DDL scientifically enables several new research paradigms: \textbf{(1) From Binary to Localized Security}: Previous datasets only support "fake vs. real" classification. DDL enables models to learn "where" and "when" the forgery happens, which is scientific crucial for interpretable forensic AI. \textbf{(2) Cross-Modal Joint Defense}: By integrating audio, visual, and multi-face scenarios in one benchmark, DDL enables research on multimodal consistency modeling. \textbf{(3) Systematic Generalization Evaluation}: Our OOD split allows researchers to systematically study the generalization boundary of detectors against entirely unseen generation architectures.
\section{Related Work}
\label{sec:realted_work}

\subsection{Deepfake Detection Datasets}
Existing deepfake detection datasets mainly offer binary labels with critical limitations across multiple aspects. Early datasets like FF++ \cite{rossler2019faceforensics++} and Celeb-DF \cite{li2020celeb} face limitations including restricted manipulation scenarios, limited deepfake diversity, and small scales. Subsequent DFDC \cite{dolhansky2020deepfake} expands scale/diversity with 8 deepfake techniques. FakeAVCeleb \cite{khalid2021fakeavceleb} introduces audio-visual forgeries. However, these datasets \cite{rossler2019faceforensics++,li2020celeb,dolhansky2020deepfake,khalid2021fakeavceleb} lack integration of diffusion-based generation techniques. While DeepFakeFace \cite{song2023robustness} and DiffusionDeepfake \cite{bhattacharyya2024diffusion} adopt advanced diffusion models, their manipulations are limited to full-face synthesis. DF40 \cite{yandf40} increases technical diversity but retains single-face constraints, omitting multi-face scenarios and complex audio-visual modalities. The absence of forgery localization annotations limits detection models to binary outputs, hindering fine-grained analysis and forensic interpretability. While some studies \cite{miao2025multi,dang2020detection,kong2022detect} attempt to generate pseudo-ground-truth masks in FF++ \cite{rossler2019faceforensics++}, annotation quality remains insufficient for high-precision interpretability. 
DDL expands forgery methods/modes/scenarios/modalities and provides accurate, fine-grained localization labels.

\subsection{Deepfake Localization Datasets}
Deepfake localization tasks and datasets have recently gained attention. ForgeryNet \cite{he2021forgerynet} provides spatial and temporal forgery labels for 15 methods but restricts to single-face scenarios with full-face manipulations. Dolos \cite{tantaru2024dolos} annotates local face manipulation masks in single-face scenarios. OpenForensics \cite{le2021openforensics} generates 100K multi-face images (3 methods) but lacks pristine image pairs. FFIW \cite{zhou2021face} extends to multi-face videos but face scale and deepfake methods are limitations. LAV-DF \cite{cai2022you} introduces cross-modal temporal annotations but contains only ~10K samples with single-modality manipulations. AV-Deepfake1M (AV-DF1M) \cite{cai2024av} scales to 0.8M samples but relies on single video/audio methods (2 categories). 
These datasets exhibit critical deficiencies in manipulation diversity, failing to address complex real-world scenarios compared to our DDL.

\section{DDL Dataset}
The DDL dataset comprises over 1.4M forged samples from 80 deepfake methods (Table \ref{tab:datasets_1}), providing fine-grained spatial and temporal annotations to support deepfake detection and localization. It is partitioned into an image subset (DDL-I) and an audio-visual subset (DDL-AV). DDL-I contains over 1.18M forged single- and multi-face images, covering traditional forgery modes—Face-Swapping (FS), Face-Reenactment (FR), Full-Face Synthesis (FFS), and Face Editing (FE)—and a novel Hybrid Face Forgery (HFF) mode. Correspondingly, DDL-AV includes over 0.23M forged audio-visual samples, encompassing traditional manipulations—Replacement (RE), Deletion (DE), and Insertion (IS)—and two novel forgery modes: Audio-Visual Asynchronous Manipulation (AVAM) and Audio-Visual Full Synthesis (AVFS).

\subsection{Source Data Collection}
\subsubsection{\textbf{Source Images.}}
We construct our dataset from 14 existing facial datasets to ensure diversity in identity, pose, expression, and scene contexts. For single-face scenarios, we use real-face datasets (FFHQ \cite{karras2019style}, CelebA \cite{liu2015deep}, CelebV-HQ \cite{zhu2022celebv}, VFHQ \cite{xie2022vfhq}, and VoxCeleb2 \cite{chung2018voxceleb2}) and the real subsets of deepfake benchmarks (FF++ \cite{rossler2019faceforensics++}, DFDC \cite{dolhansky2020deepfake}, and Celeb-DF \cite{li2020celeb}). Similarly, multi-face data is sourced from real-face datasets (WiderFace \cite{yang2016wider} and FDDB \cite{jain2010fddb}), general-purpose datasets (OpenImage \cite{kuznetsova2020open} and VisualNews \cite{liu2020visual}), and deepfake datasets (FFIW \cite{zhou2021face} and Manual-Fake \cite{haiwei2022exploring}).
To construct controlled distribution shifts and prevent source overlap, these sources are partitioned between our training and testing splits based on their size and original purpose, as detailed in Table \ref{tab:datasets_2}.

During preprocessing, we use MTCNN to extract facial regions from single-face images. For multi-face images, RetinaFace is employed to select those containing at least two distinct identities. From videos, we sample a single frame to reduce redundancy and enhance identity diversity.

\subsubsection{\textbf{Source Audio-Video Data.}}
We leverage two existing real audio-video face datasets, VoxCeleb2 \cite{chung2018voxceleb2} and CelebV-HQ \cite{zhu2022celebv}, as our original data sources. For the training and validation sets, we utilize the `dev' subset of   VoxCeleb2. The test set comprises the CelebV-HQ dataset and `test' subset of  VoxCeleb2, with a selection strategy designed to maximize the distributional discrepancy between the training and test sets. 

During the data preprocessing phase, our video preprocessing includes random single-frame extraction, random video clip selection, and full-length video utilization. For audio modality, we employ speech content extraction and recognition techniques, as well as foreground-background audio separation. The preprocessed audio-visual materials are flexibly selected based on the specific requirements of the generative methods, ensuring adaptability to various model training and testing needs.

\begin{table}[tp]
\caption{Statistics of original data sources in the DDL dataset.}
\centering
\scalebox{0.60}{
\begin{tabular}{c|c|c|c}
\toprule
Types & Train \& Valid & Test & Total \\ \midrule
Image & \begin{tabular}[c]{@{}c@{}}FF++ (1K), CelebDF (1K), Manual-Fake (2K),FFIW (10K), \\ FDDB (11K), DFDC (19K), FFHQ (15K), WiderFace (16K), \\ CelebA (30K),CelebV-HQ (30K), Open Image (60K)\end{tabular} & \begin{tabular}[c]{@{}c@{}}VoxCeleb2 (7K),\\ VFHQ (15K),\\ VisualNews (36K)\end{tabular} & 253K \\ \midrule
Audio-Video & VoxCeleb2 (73K) & \begin{tabular}[c]{@{}c@{}}VoxCeleb2 (10K),\\ CelebV-HQ (16K)\end{tabular} & 99K \\ \midrule
Total & 268K & 84K & 352K \\ \bottomrule
\end{tabular}
}
\label{tab:datasets_2}
\end{table}

\subsection{Deepfake Generation} 
According to the modality of the input data (e.g., image, audio, video) and the scene configuration (e.g., single-face, multi-face, audio–video), we design four targeted deepfake generation pipelines. The detailed generation procedures are described as follows (the complete workflow diagrams are provided on GitHub).

\subsubsection{\textbf{Single-Face Scene Manipulation}}
We propose a mask-guided pipeline for manipulating single-face images. The process begins by employing a face parsing module to extract landmarks, which guide the generation of random masks over local regions (e.g., eyes, nose). A generative model is then conditioned on the original image, the mask, and a target attribute prompt to selectively alter attributes like expression or age.
This mechanism ensures modifications are localized, thereby preserving the subject's identity and background. The pipeline outputs both the manipulated image and a corresponding mask that precisely delineates the forged pixels. 
Our pipeline employs both GAN-based and diffusion-based generative models, including FaceInpainting \cite{luo2023reference}, GLIDE \cite{nichol2021glide}, and PILOT \cite{pan2024coherent}. Additional models are detailed on GitHub.

\subsubsection{\textbf{Multi-Face Scene Manipulation}}
We propose a pipeline for multi-face manipulation in images. A detection module first localizes all faces and generates corresponding binary masks. For each target, a source identity is randomly sampled from a database, and a generative model performs identity replacement on the masked region via predefined strategies (e.g., partial or full swapping). A blending module then seamlessly integrates the synthesized faces, preserving consistency with the background and unmodified subjects. 
For this scenario, we expand our scope beyond GAN and Diffusion frameworks to include novel architectures such as NeRF, Autoregressive, Gaussian Splatting, Flow, and VAE, with representative methods like LivePortrait \cite{guo2024liveportrait}, FaceDancer \cite{rosberg2023facedancer}, ICLight \cite{zhang2025scaling}, AD-NeRF \cite{guo2021ad}, ControlAR \cite{li2024controlar}, GaussianEditor \cite{chen2024gaussianeditor}, StyleFlow \cite{abdal2021styleflow}, and SQ-VAE \cite{takida2022sq}. 
A comprehensive list is available on GitHub.

\subsubsection{\textbf{Audio–Video Full-Synthesis Manipulation}}
This pipeline synthesizes full audio-video content from text (or reference audio) and a target identity. An audio-video model generates a facial sequence conditioned on the speaker's identity features and driven by a corresponding speech signal synthesized from the text. This process ensures that lip movements, expressions, and head poses are synchronized with the audio's acoustic properties, resulting in a complete deepfake video. 
To accomplish this, we employ joint audio-video models (e.g., Kling2.1 \cite{kling2025}, Jimeng3.0 \cite{jimeng2025}) as well as pipelines combining video generation (e.g., Pika1.5 \cite{pika2025}, MAGI-1 \cite{ai2025magi1autoregressivevideogeneration}) with video-to-audio models (e.g., AudioX \cite{tian2025audiox}, MMAudio \cite{cheng2025taming}), with full details on GitHub.

\subsubsection{\textbf{Audio–Video Local Manipulation}}
This pipeline performs fine-grained local manipulation of existing audio-video content. Given an original video and a driving signal (e.g., new text), the system first transcribes the original audio. It then edits the text (e.g., word substitution, deletion) while recording the temporal locations (start/end timestamps) of the modifications. Subsequently, a new, denoised speech signal is synthesized from the edited text. An audio-driven model then re-renders the lip region via geometric deformation and texture synthesis to ensure precise synchronization with the new audio, while preserving the subject's identity and background. The pipeline yields two outputs: (1) the manipulated video with alterations confined to the lip region, and (2) temporal segment annotations that delineate the forged intervals. 
In practice, in addition to audio-driven local facial generation methods (e.g., AniPortrait \cite{wei2024aniportrait}, VideoRetalking \cite{cheng2022videoretalking}), we also employ face swapping techniques (e.g., InSwapper \cite{inswapper2023}, SimSwap \cite{chen2020simswap}) to enhance data diversity. Further details about these models are available on GitHub.

\subsection{Real-world Perturbations}
Forged samples are subject to various perturbations during real-world transmission. To simulate this scenario, we design and apply 27 types of perturbation methods to the test set samples. Specifically, for the image modality, perturbations are categorized into three groups: color, corruption, and weather, with each category comprising 8 distinct perturbation methods. For the audio-visual modality, we employ a joint perturbation strategy, including H.264-based compression, Gaussian noise, and reverberation blur. Representative examples can be found on GitHub.

\subsection{Data Partitioning}
We partition the dataset into train, valid, and test sets, as shown in Table \ref{tab:dataset_analysis_2}. The validation set is randomly held out from the training pool and excluded from model training. Real samples are partitioned based on their original dataset sources, while fake samples are classified according to different types of generation methods applied to the real data. Detailed partitioning protocols are provided on GitHub.

\begin{table}[tp]
\caption{Statistics of DDL datasets.}
\centering
\scalebox{0.88}{
\begin{tabular}{l|cc|cc|cc|cc}
\toprule
\multirow{2}{*}{Subsets} & \multicolumn{2}{c|}{Train} & \multicolumn{2}{c|}{Valid} & \multicolumn{2}{c|}{Test} & \multicolumn{2}{c}{Total} \\ \cmidrule(r){2-3} \cmidrule(r){4-6} \cmidrule(r){7-9}
 & Real & Fake & Real & Fake & Real & Fake & Real & Fake \\ 
 \cmidrule(r){1-1} \cmidrule(r){2-2} \cmidrule(r){3-3} \cmidrule(r){4-4} \cmidrule(r){5-5} \cmidrule(r){6-6} \cmidrule(r){7-7} \cmidrule(r){8-8} \cmidrule(r){9-9}
DDL-I & 156K & 799K & 39K & 199K & 58K & 182K & 253K & 1180K \\
DDL-AV & 68K & 134K & 5K & 9K & 26K & 88K & 99K & 231K \\ \bottomrule
\end{tabular}
}
\Description{}
\label{tab:dataset_analysis_2}
\end{table}

\begin{table}[tp]
\caption{Statistics of forgery modes in deepfake datasets.}
\centering
\scalebox{0.80}{
\begin{tabular}{lcccccccccc}
\toprule
Datasets & FS & FR & FFS & FE & HFF & DE & RE & IS & AVAM & AVFS \\ \midrule
\textbf{DDL (ours)}         & $\checkmark$ & $\checkmark$ & $\checkmark$ & $\checkmark$ & $\checkmark$ & $\checkmark$ & $\checkmark$ & $\checkmark$ & $\checkmark$ & $\checkmark$ \\
DF40        & $\checkmark$ & $\checkmark$ & $\checkmark$ & $\checkmark$ & $\times$     & $\times$     & $\times$     & $\times$     & $\times$     & $\times$     \\
AV-DF1M     & $\times$     & $\times$     & $\times$     & $\times$     & $\times$     & $\checkmark$ & $\checkmark$ & $\checkmark$ & $\times$ & $\times$     \\
ForgeryNet  & $\checkmark$ & $\checkmark$ & $\times$     & $\checkmark$ & $\times$ & $\times$     & $\times$     & $\times$     & $\times$     & $\times$     \\ \bottomrule
\end{tabular}
}
\Description{}
\label{tab:dataset_analysis_1}
\end{table}

\begin{figure}[t]
    \centering
    \includegraphics[width=0.99\linewidth]{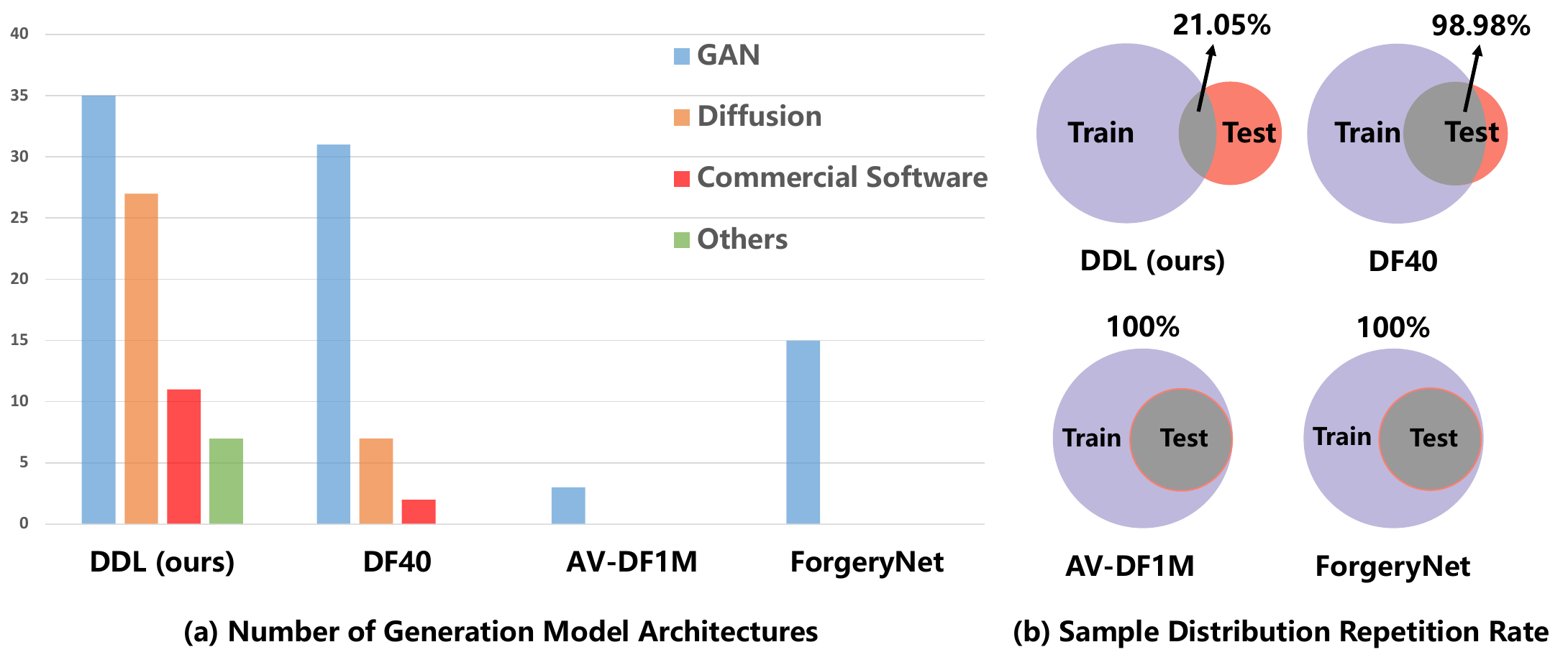}
    \caption{Dataset statistical analysis. (a) Number of deepfake methods under different generation model architectures. "Others" denotes the total of VAE, NeRF, Normalizing Flow, and Autoregressive. (b) Proportion of test set samples originating from the same source as the training set. 
    }
    \label{fig:dataset_analysis_1}
\end{figure}

\subsection{Dataset Characteristics}

\subsubsection{\textbf{Diverse Forgery Scenarios and Modalities.}}
As shown in Table \ref{tab:datasets_1}, `Scenarios' and `Modality' columns, our DDL dataset encompasses common real-world scenarios, including single-face, multi-face, and audio-visual, while also incorporating generation models across three modalities: audio, image, and video. Compared to the DF40 \cite{yandf40} and AV-DF1M \cite{cai2024av} datasets, DDL effectively addresses the limitations of existing datasets by offering greater diversity in forgery scenarios and richer coverage of modality.

\subsubsection{\textbf{Varied Manipulation Modes.}}
As shown in Table \ref{tab:dataset_analysis_1}, our DDL dataset covers a full range of traditional image forgery modes (FS, FR, FFS, FE) and audio-video modes (RE, DE, IS). Building on new AIGC methods and real-world needs, we further introduce novel forgery types such as HFF, AVFS, and AVAM (Detailed visual examples are provided on GitHub). Compared to the state-of-the-art image DF40 \cite{yandf40} dataset, DDL adds 6 new forgery categories, and compared to the latest audio-video AV-DF1M \cite{cai2024av} dataset, it introduces 7 additional modes.

\subsubsection{\textbf{Comprehensive Generation Model Architectures and Types.}}
Unlike previous datasets that primarily focus on generated samples based on GAN or Diffusion models, the proposed DDL dataset fully accounts for the diversity of generative model architectures encountered in real-world scenarios. Specifically, we collect seven representative categories of generative models, including 35 GANs, 27 Diffusion models, 3 VAEs, 2 Normalizing Flows, 1 Autoregressive model, 1 NeRF, as well as 11 popular commercial software tools. As shown in Fig. \ref{fig:dataset_analysis_1}(a), DDL offers more comprehensive coverage of architectural diversity and model types in generative models compared to existing datasets.

\subsubsection{\textbf{Out-of-Distribution Test Set.}}
In real-world scenarios, the distribution of forged data often differs significantly from that of the training set. However, most existing datasets are constructed such that the training and test sets share the same distribution, which can easily lead to model overfitting during training and impede the model's ability to effectively handle unseen conditions in real-world applications. Therefore, our DDL dataset deliberately isolates the distributions of the training and test sets from two perspectives: the sources of real data and the generation model types used. This enables the construction of out-of-distribution test sets. As shown in Fig. \ref{fig:dataset_analysis_1}(b), the source overlap rate of test samples in DDL is only 21.05\%, compared to 98.98\% for DF40 \cite{yandf40}, and a full 100\% overlap for both AV-DF1M \cite{cai2024av} and ForgeryNet \cite{he2021forgerynet}.

\section{Evaluations}

\begin{table}[tp]
  \caption{Benchmark Experiments on DDL datasets.}
  \label{tab:combined_benchmark}
  \centering
  \resizebox{\columnwidth}{!}{%
  \begin{tabular}{@{}l|cccc|cccc@{}}
    \toprule
    Models & \multicolumn{4}{c|}{Valid Set} & \multicolumn{4}{c}{Test Set} \\
    \midrule
    \multicolumn{1}{l|}{\textit{DDL-I Dataset}} & ACC & AUC & F1 & IoU & ACC & AUC & F1 & IoU \\
    \cmidrule(lr){2-5} \cmidrule(lr){6-9}
    UnivFD \cite{ojha2023towards} & 83.64 & 71.50 & -- & -- & 76.12 & 70.14 & -- & -- \\
    DualNet \cite{xi2023ai} & 80.97 & 83.21 & -- & -- & 74.65 & 65.96 & -- & -- \\
    SPSL \cite{liu2021spatial} & 81.34 & 79.27 & -- & -- & 72.79 & 65.65 & -- & -- \\
    IML-ViT \cite{ma2023iml} & 90.39 & 95.07 & 73.45 & 69.28 & 80.35 & 86.52 & 45.77 & 41.93 \\
    MSCCNet \cite{miao2025multi} & 99.83 & 99.95 & 98.14 & 96.79 & 78.17 & 88.88 & 66.39 & 49.69 \\
    SparseViT \cite{su2025can} & 99.70 & 99.98 & 97.93 & 95.94 & 85.47 & 92.53 & 69.40 & 53.13 \\
    Mesorch \cite{zhu2025mesoscopic} & 99.48 & 99.98 & 98.25 & 96.56 & 80.15 & 89.17 & 52.43 & 35.52 \\
    \midrule
    \multicolumn{1}{l|}{\textit{DDL-AV Dataset}} & ACC & AUC & AP & AR & ACC & AUC & AP & AR \\
    \cmidrule(lr){2-5} \cmidrule(lr){6-9}
    BA-TFD \cite{cai2022you} & 67.68 & 95.23 & 47.20 & 36.17 & 42.71 & 68.60 & 26.35 & 23.53 \\
    BA-TFD+ \cite{cai2023glitch} & 70.71 & 94.90 & 53.25 & 40.45 & 42.74 & 54.40 & 28.10 & 24.72 \\
    UMMAFormer \cite{zhang2023ummaformer} & 78.87 & 93.35 & 72.28 & 58.68 & 54.39 & 60.07 & 55.75 & 56.14 \\
    \bottomrule
  \end{tabular}%
  }
\end{table}

\subsection{Experiment Settings}

\subsubsection{\textbf{Baseline Models.}}
Considering the requirement for simultaneous forgery detection and localization tasks, we selected models with publicly reproducible code: 
(1) For deepfake image detection, we use UnivFD \cite{ojha2023towards}, DualNet \cite{xi2023ai}, and SPSL \cite{liu2021spatial}; 
(2) For deepfake image detection and localization, we employ IML-ViT \cite{ma2023iml}, MSCCNet \cite{miao2025multi}, SparseViT \cite{su2025can}, and Mesorch \cite{zhu2025mesoscopic}; 
(3) For deepfake audio-visual detection and localization, we utilize BA-TFD \cite{cai2022you}, BA-TFD+ \cite{cai2023glitch}, and UMMAFormer \cite{zhang2023ummaformer}.

\subsubsection{\textbf{Evaluation Metrics.}}
For detection tasks, we adopt widely used ACC and AUC as evaluation standards. 
For localization tasks, spatial localization is evaluated using F1 and IoU, whereas temporal localization is evaluated using AP at a temporal IoU threshold of 0.5 and AR with at most five proposals per sample. Localization metrics are computed only for forged samples.

\subsection{Experiments and Analysis}

\subsubsection{\textbf{Evaluation on DDL-I Dataset.}}
We conduct experiments to analyze the credibility and effectiveness of the DDL-I dataset in assessing model generalization and robustness capabilities, with a focus on localization tasks.

\noindent \textbf{Generalization:}
As shown in Table \ref{tab:combined_benchmark}, we evaluate models on the DDL-I dataset. All models perform well on the validation set, demonstrating strong classification and localization abilities. However, performance notably drops on the test set, especially in localization. For example, Mesorch's IoU falls to 35.52\%, highlighting a significant distribution shift between test and training, which aligns with the unknown characteristics of forged samples in real-world scenarios.

\begin{figure}[tp]
    \centering
    \includegraphics[width=0.99\linewidth]{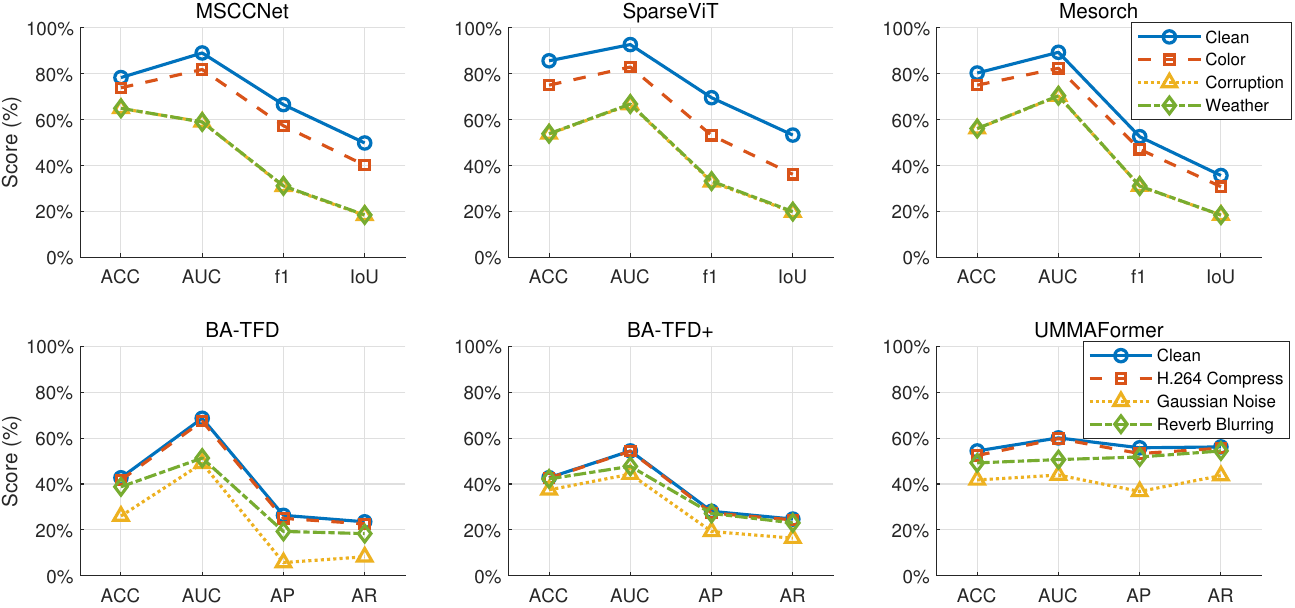}
    \caption{Robustness experiments of DDL-I (top) and DDL-AV (bottom) datasets. Clean represents the raw test set.
    }
    \Description{}
    \label{fig:robustness}
\end{figure}

\noindent \textbf{Robustness:}
Fig. \ref{fig:robustness} shows that all models degrade under various perturbations. Color perturbations have minor effects, while Corruption and Weather cause greater performance drops. Notably, SparseViT's IoU decreases from 53.13\% to 19.81\% under Weather. These results demonstrate that our perturbations capture common real-world perturbations, offering valuable guidance for data augmentation in future model training.

\subsubsection{\textbf{Evaluation on DDL-AV Dataset.}}
We design experiments to evaluate the credibility and effectiveness of the proposed DDL-AV dataset in assessing model generalization capabilities and robustness, with a focus on localization tasks. 

\noindent \textbf{Generalization:}
As shown in Table \ref{tab:combined_benchmark}, existing models achieve superior video-level classification performance and moderate localization results on validation sets, indicating the inherent difficulty of temporal localization. However, our designed out-of-distribution test sets cause significant performance degradation, with average classification AUC drops of 33.47 percentage points and a decrease of 15.73 percentage points (38.9\% relative) in localization AR for BA-TFD+.

\noindent \textbf{Robustness:}
As illustrated in Fig. \ref{fig:robustness}, all models demonstrate performance degradation under audio-visual perturbations, with Gaussian noise perturbations causing the most severe degradation. The BA-TFD model shows the most significant localization performance decline, while UMMAFormer exhibits relatively optimal robustness. These results validate DDL-AV's ability to reveal model vulnerabilities under realistic interference.
\section{Conclusion}
This paper introduces a large-scale, diverse, multi-modal, and multi-scenario DDL dataset, which contains 1.4M+ forged samples generated using 80 different deepfake methods. The DDL dataset provides rich spatial and temporal interpretability annotations, supporting the transition of deepfake detection tasks towards localization and interpretability, thereby enabling potential applications in rigorous domains such as legal proceedings. Extensive benchmark experiments demonstrate that the test sets in the DDL dataset, which conform to real-world data distributions, will effectively facilitate further research in next-generation deepfake detection, localization, and interpretability tasks.

\section{Broader Impact and Ethics Statement} 
DDL establishes a foundational benchmark for deepfake detection and localization via its large-scale, diverse, multi-modal data. To ensure ethical compliance, all source data is licensed for reuse and privacy-compliant, while usage agreements restrict access to research purposes, mitigating potential risks.

\begin{acks}
This work was supported by the Ant Group Postdoctoral Programme.
This work was also supported by the National Research Foundation Singapore, under its AI Singapore Programme (AISG Award No: AISG3-RP-2024-033).
\end{acks}


\bibliographystyle{ACM-Reference-Format}
\balance
\bibliography{sample-base}

@inproceedings{rossler2019faceforensics++,
  title={Faceforensics++: Learning to detect manipulated facial images},
  author={Rossler, Andreas and Cozzolino, Davide and Verdoliva, Luisa and Riess, Christian and Thies, Justus and Nie{\ss}ner, Matthias},
  booktitle={Proceedings of the IEEE/CVF international conference on computer vision},
  pages={1--11},
  year={2019}
}

@article{dolhansky2020deepfake,
  title={The deepfake detection challenge (dfdc) dataset},
  author={Dolhansky, Brian and Bitton, Joanna and Pflaum, Ben and Lu, Jikuo and Howes, Russ and Wang, Menglin and Ferrer, Cristian Canton},
  journal={arXiv preprint arXiv:2006.07397},
  year={2020}
}

@inproceedings{yang2016wider,
  title={Wider face: A face detection benchmark},
  author={Yang, Shuo and Luo, Ping and Loy, Chen-Change and Tang, Xiaoou},
  booktitle={Proceedings of the IEEE conference on computer vision and pattern recognition},
  pages={5525--5533},
  year={2016}
}

@techreport{jain2010fddb,
  title={Fddb: A benchmark for face detection in unconstrained settings},
  author={Jain, Vidit and Learned-Miller, Erik},
  year={2010},
  institution={UMass Amherst technical report}
}

@article{liu2020visual,
  title={Visual news: Benchmark and challenges in news image captioning},
  author={Liu, Fuxiao and Wang, Yinghan and Wang, Tianlu and Ordonez, Vicente},
  journal={arXiv preprint arXiv:2010.03743},
  year={2020}
}

@inproceedings{guo2021ad,
  title={Ad-nerf: Audio driven neural radiance fields for talking head synthesis},
  author={Guo, Yudong and Chen, Keyu and Liang, Sen and Liu, Yong-Jin and Bao, Hujun and Zhang, Juyong},
  booktitle={Proceedings of the IEEE/CVF international conference on computer vision},
  pages={5784--5794},
  year={2021}
}

@article{tian2024visual,
  title={Visual autoregressive modeling: Scalable image generation via next-scale prediction},
  author={Tian, Keyu and Jiang, Yi and Yuan, Zehuan and Peng, Bingyue and Wang, Liwei},
  journal={arXiv preprint arXiv:2404.02905},
  year={2024}
}

@article{guo2024liveportrait,
  title={Liveportrait: Efficient portrait animation with stitching and retargeting control},
  author={Guo, Jianzhu and Zhang, Dingyun and Liu, Xiaoqiang and Zhong, Zhizhou and Zhang, Yuan and Wan, Pengfei and Zhang, Di},
  journal={arXiv preprint arXiv:2407.03168},
  year={2024}
}

@inproceedings{dang2020detection,
  title={On the detection of digital face manipulation},
  author={Dang, Hao and Liu, Feng and Stehouwer, Joel and Liu, Xiaoming and Jain, Anil K},
  booktitle={Proceedings of the IEEE/CVF Conference on Computer Vision and Pattern recognition},
  pages={5781--5790},
  year={2020}
}

@inproceedings{he2021forgerynet,
  title={Forgerynet: A versatile benchmark for comprehensive forgery analysis},
  author={He, Yinan and Gan, Bei and Chen, Siyu and Zhou, Yichun and Yin, Guojun and Song, Luchuan and Sheng, Lu and Shao, Jing and Liu, Ziwei},
  booktitle={Proceedings of the IEEE/CVF conference on computer vision and pattern recognition},
  pages={4360--4369},
  year={2021}
}

@article{song2023robustness,
  title={Robustness and generalizability of deepfake detection: A study with diffusion models},
  author={Song, Haixu and Huang, Shiyu and Dong, Yinpeng and Tu, Wei-Wei},
  journal={arXiv preprint arXiv:2309.02218},
  year={2023}
}

@article{bhattacharyya2024diffusion,
  title={Diffusion deepfake},
  author={Bhattacharyya, Chaitali and Wang, Hanxiao and Zhang, Feng and Kim, Sungho and Zhu, Xiatian},
  journal={arXiv preprint arXiv:2404.01579},
  year={2024}
}

@article{khalid2021fakeavceleb,
  title={FakeAVCeleb: A novel audio-video multimodal deepfake dataset},
  author={Khalid, Hasam and Tariq, Shahroz and Kim, Minha and Woo, Simon S},
  journal={arXiv preprint arXiv:2108.05080},
  year={2021}
}

@inproceedings{cai2022you,
  title={Do you really mean that? content driven audio-visual deepfake dataset and multimodal method for temporal forgery localization},
  author={Cai, Zhixi and Stefanov, Kalin and Dhall, Abhinav and Hayat, Munawar},
  booktitle={2022 International Conference on Digital Image Computing: Techniques and Applications (DICTA)},
  pages={1--10},
  year={2022},
  organization={IEEE}
}

@inproceedings{cai2024av,
  title={AV-Deepfake1M: A large-scale LLM-driven audio-visual deepfake dataset},
  author={Cai, Zhixi and Ghosh, Shreya and Adatia, Aman Pankaj and Hayat, Munawar and Dhall, Abhinav and Gedeon, Tom and Stefanov, Kalin},
  booktitle={Proceedings of the 32nd ACM International Conference on Multimedia},
  pages={7414--7423},
  year={2024}
}

@inproceedings{rosberg2023facedancer,
  title={Facedancer: Pose-and occlusion-aware high fidelity face swapping},
  author={Rosberg, Felix and Aksoy, Eren Erdal and Alonso-Fernandez, Fernando and Englund, Cristofer},
  booktitle={Proceedings of the IEEE/CVF winter conference on applications of computer vision},
  pages={3454--3463},
  year={2023}
}

@inproceedings{zhou2021face,
  title={Face forensics in the wild},
  author={Zhou, Tianfei and Wang, Wenguan and Liang, Zhiyuan and Shen, Jianbing},
  booktitle={Proceedings of the IEEE/CVF conference on computer vision and pattern recognition},
  pages={5778--5788},
  year={2021}
}

@inproceedings{le2021openforensics,
  title={Openforensics: Large-scale challenging dataset for multi-face forgery detection and segmentation in-the-wild},
  author={Le, Trung-Nghia and Nguyen, Huy H and Yamagishi, Junichi and Echizen, Isao},
  booktitle={Proceedings of the IEEE/CVF International Conference on Computer Vision},
  pages={10117--10127},
  year={2021}
}

@article{zhang2024comics,
  author={Zhang, Cong and Qi, Honggang and Wang, Shuhui and Li, Yuezun and Lyu, Siwei},
  journal={IEEE Transactions on Circuits and Systems for Video Technology}, 
  title={COMICS: End-to-end Bi-grained Contrastive Learning for Multi-face Forgery Detection}, 
  year={2024},
  publisher={IEEE}
}

@article{kong2022detect,
  title={Detect and locate: Exposing face manipulation by semantic-and noise-level telltales},
  author={Kong, Chenqi and Chen, Baoliang and Li, Haoliang and Wang, Shiqi and Rocha, Anderson and Kwong, Sam},
  journal={IEEE Transactions on Information Forensics and Security},
  volume={17},
  pages={1741--1756},
  year={2022},
  publisher={IEEE}
}

@article{huang2022fakelocator,
  title={FakeLocator: Robust localization of GAN-based face manipulations},
  author={Huang, Yihao and Juefei-Xu, Felix and Guo, Qing and Liu, Yang and Pu, Geguang},
  journal={IEEE Transactions on Information Forensics and Security},
  volume={17},
  pages={2657--2672},
  year={2022},
  publisher={IEEE}
}

@inproceedings{yan2023ucf,
  title={Ucf: Uncovering common features for generalizable deepfake detection},
  author={Yan, Zhiyuan and Zhang, Yong and Fan, Yanbo and Wu, Baoyuan},
  booktitle={Proceedings of the IEEE/CVF International Conference on Computer Vision},
  pages={22412--22423},
  year={2023}
}

@inproceedings{oorloff2024avff,
  title={AVFF: Audio-Visual Feature Fusion for Video Deepfake Detection},
  author={Oorloff, Trevine and Koppisetti, Surya and Bonettini, Nicol{\`o} and Solanki, Divyaraj and Colman, Ben and Yacoob, Yaser and Shahriyari, Ali and Bharaj, Gaurav},
  booktitle={Proceedings of the IEEE/CVF Conference on Computer Vision and Pattern Recognition},
  pages={27102--27112},
  year={2024}
}

@inproceedings{lin2024fake,
  title={Fake it till you make it: Curricular dynamic forgery augmentations towards general deepfake detection},
  author={Lin, Yuzhen and Song, Wentang and Li, Bin and Li, Yuezun and Ni, Jiangqun and Chen, Han and Li, Qiushi},
  booktitle={European Conference on Computer Vision},
  pages={104--122},
  year={2024},
  organization={Springer}
}

@article{sun2024diffusionfake,
  title={DiffusionFake: Enhancing Generalization in Deepfake Detection via Guided Stable Diffusion},
  author={Sun, Ke and Chen, Shen and Yao, Taiping and Liu, Hong and Sun, Xiaoshuai and Ding, Shouhong and Ji, Rongrong},
  journal={arXiv preprint arXiv:2410.04372},
  year={2024}
}

@inproceedings{li2020celeb,
  title={Celeb-df: A large-scale challenging dataset for deepfake forensics},
  author={Li, Yuezun and Yang, Xin and Sun, Pu and Qi, Honggang and Lyu, Siwei},
  booktitle={Proceedings of the IEEE/CVF conference on computer vision and pattern recognition},
  pages={3207--3216},
  year={2020}
}

@article{jiang2021deeperforensics,
  title={DeeperForensics Challenge 2020 on real-world face forgery detection: Methods and results},
  author={Jiang, Liming and Guo, Zhengkui and Wu, Wayne and Liu, Zhaoyang and Liu, Ziwei and Loy, Chen Change and Yang, Shuo and Xiong, Yuanjun and Xia, Wei and Chen, Baoying and others},
  journal={arXiv preprint arXiv:2102.09471},
  year={2021}
}

@article{luo2023reference,
  title={Reference-guided large-scale face inpainting with identity and texture control},
  author={Luo, Wuyang and Yang, Su and Zhang, Weishan},
  journal={IEEE Transactions on Circuits and Systems for Video Technology},
  volume={33},
  number={10},
  pages={5498--5509},
  year={2023},
  publisher={IEEE}
}

@article{nichol2021glide,
  title={Glide: Towards photorealistic image generation and editing with text-guided diffusion models},
  author={Nichol, Alex and Dhariwal, Prafulla and Ramesh, Aditya and Shyam, Pranav and Mishkin, Pamela and McGrew, Bob and Sutskever, Ilya and Chen, Mark},
  journal={arXiv preprint arXiv:2112.10741},
  year={2021}
}

@article{pan2024coherent,
  title={Coherent and Multi-modality Image Inpainting via Latent Space Optimization},
  author={Pan, Lingzhi and Zhang, Tong and Chen, Bingyuan and Zhou, Qi and Ke, Wei and S{\"u}sstrunk, Sabine and Salzmann, Mathieu},
  journal={arXiv preprint arXiv:2407.08019},
  year={2024}
}

@inproceedings{chen2020simswap,
  title={Simswap: An efficient framework for high fidelity face swapping},
  author={Chen, Renwang and Chen, Xuanhong and Ni, Bingbing and Ge, Yanhao},
  booktitle={Proceedings of the 28th ACM international conference on multimedia},
  pages={2003--2011},
  year={2020}
}

@article{wei2024aniportrait,
  title={Aniportrait: Audio-driven synthesis of photorealistic portrait animation},
  author={Wei, Huawei and Yang, Zejun and Wang, Zhisheng},
  journal={arXiv preprint arXiv:2403.17694},
  year={2024}
}

@inproceedings{zhang2025scaling,
  title={Scaling in-the-wild training for diffusion-based illumination harmonization and editing by imposing consistent light transport},
  author={Zhang, Lvmin and Rao, Anyi and Agrawala, Maneesh},
  booktitle={The Thirteenth International Conference on Learning Representations},
  year={2025}
}

@article{kuckreja2025tell,
  title={Tell me Habibi, is it Real or Fake?},
  author={Kuckreja, Kartik and Gupta, Parul and Hamed, Injy and Solorio, Thamar and Khan, Muhammad Haris and Dhall, Abhinav},
  journal={arXiv preprint arXiv:2505.22581},
  year={2025}
}

@article{li2024controlar,
  title={Controlar: Controllable image generation with autoregressive models},
  author={Li, Zongming and Cheng, Tianheng and Chen, Shoufa and Sun, Peize and Shen, Haocheng and Ran, Longjin and Chen, Xiaoxin and Liu, Wenyu and Wang, Xinggang},
  journal={arXiv preprint arXiv:2410.02705},
  year={2024}
}

@inproceedings{chen2024gaussianeditor,
  title={Gaussianeditor: Swift and controllable 3d editing with gaussian splatting},
  author={Chen, Yiwen and Chen, Zilong and Zhang, Chi and Wang, Feng and Yang, Xiaofeng and Wang, Yikai and Cai, Zhongang and Yang, Lei and Liu, Huaping and Lin, Guosheng},
  booktitle={Proceedings of the IEEE/CVF Conference on Computer Vision and Pattern Recognition},
  pages={21476--21485},
  year={2024}
}

@article{abdal2021styleflow,
  title={Styleflow: Attribute-conditioned exploration of stylegan-generated images using conditional continuous normalizing flows},
  author={Abdal, Rameen and Zhu, Peihao and Mitra, Niloy J and Wonka, Peter},
  journal={ACM Transactions on Graphics (ToG)},
  volume={40},
  number={3},
  pages={1--21},
  year={2021},
  publisher={ACM New York, NY}
}

@article{takida2022sq,
  title={Sq-vae: Variational bayes on discrete representation with self-annealed stochastic quantization},
  author={Takida, Yuhta and Shibuya, Takashi and Liao, WeiHsiang and Lai, Chieh-Hsin and Ohmura, Junki and Uesaka, Toshimitsu and Murata, Naoki and Takahashi, Shusuke and Kumakura, Toshiyuki and Mitsufuji, Yuki},
  journal={arXiv preprint arXiv:2205.07547},
  year={2022}
}

@misc{inswapper2023,
 author={haofanwang},
 title="\url{https://github.com/haofanwang/inswapper/}",
 year={2023}
}

@inproceedings{cheng2022videoretalking,
  title={Videoretalking: Audio-based lip synchronization for talking head video editing in the wild},
  author={Cheng, Kun and Cun, Xiaodong and Zhang, Yong and Xia, Menghan and Yin, Fei and Zhu, Mingrui and Wang, Xuan and Wang, Jue and Wang, Nannan},
  booktitle={SIGGRAPH Asia 2022 Conference Papers},
  pages={1--9},
  year={2022}
}

@inproceedings{yandf40,
  title={DF40: Toward Next-Generation Deepfake Detection},
  author={Yan, Zhiyuan and Yao, Taiping and Chen, Shen and Zhao, Yandan and Fu, Xinghe and Zhu, Junwei and Luo, Donghao and Wang, Chengjie and Ding, Shouhong and Wu, Yunsheng and others},
  booktitle={The Thirty-eighth Conference on Neural Information Processing Systems Datasets and Benchmarks Track},
  year = {2024}
}

@article{goodfellow2014generative,
  title={Generative adversarial nets},
  author={Goodfellow, Ian and Pouget-Abadie, Jean and Mirza, Mehdi and Xu, Bing and Warde-Farley, David and Ozair, Sherjil and Courville, Aaron and Bengio, Yoshua},
  journal={Advances in neural information processing systems},
  volume={27},
  year={2014}
}

@article{ho2020denoising,
  title={Denoising diffusion probabilistic models},
  author={Ho, Jonathan and Jain, Ajay and Abbeel, Pieter},
  journal={Advances in neural information processing systems},
  volume={33},
  pages={6840--6851},
  year={2020}
}

@article{mildenhall2021nerf,
  title={Nerf: Representing scenes as neural radiance fields for view synthesis},
  author={Mildenhall, Ben and Srinivasan, Pratul P and Tancik, Matthew and Barron, Jonathan T and Ramamoorthi, Ravi and Ng, Ren},
  journal={Communications of the ACM},
  volume={65},
  number={1},
  pages={99--106},
  year={2021},
  publisher={ACM New York, NY, USA}
}

@inproceedings{tantaru2024dolos,
    author    = {Ț\^anțaru, Dragoș-Constantin and Oneaț\u{a}, Elisabeta and Oneaț\u{a}, Dan},
    title     = {Weakly-Supervised Deepfake Localization in Diffusion-Generated Images},
    booktitle = {Proceedings of the IEEE/CVF Winter Conference on Applications of Computer Vision (WACV)},
    year      = {2024},
    pages     = {6258-6268}
}

@article{haiwei2022exploring,
  title={Exploring spatial-temporal features for deepfake detection and localization},
  author={Haiwei, Wu and Jiantao, Zhou and Shile, Zhang and Jinyu, Tian},
  journal={arXiv preprint arXiv:2210.15872},
  year={2022}
}

@inproceedings{karras2019style,
  title={A style-based generator architecture for generative adversarial networks},
  author={Karras, Tero and Laine, Samuli and Aila, Timo},
  booktitle={Proceedings of the IEEE/CVF conference on computer vision and pattern recognition},
  pages={4401--4410},
  year={2019}
}

@inproceedings{zhu2025mesoscopic,
  title={Mesoscopic insights: orchestrating multi-scale \& hybrid architecture for image manipulation localization},
  author={Zhu, Xuekang and Ma, Xiaochen and Su, Lei and Jiang, Zhuohang and Du, Bo and Wang, Xiwen and Lei, Zeyu and Feng, Wentao and Pun, Chi-Man and Zhou, Ji-Zhe},
  booktitle={Proceedings of the AAAI Conference on Artificial Intelligence},
  pages={11022--11030},
  year={2025}
}

@inproceedings{su2025can,
  title={Can we get rid of handcrafted feature extractors? sparsevit: Nonsemantics-centered, parameter-efficient image manipulation localization through spare-coding transformer},
  author={Su, Lei and Ma, Xiaochen and Zhu, Xuekang and Niu, Chaoqun and Lei, Zeyu and Zhou, Ji-Zhe},
  booktitle={Proceedings of the AAAI Conference on Artificial Intelligence},
  pages={7024--7032},
  year={2025}
}

@inproceedings{zhang2023ummaformer,
    title={UMMAFormer: A Universal Multimodal-adaptive Transformer Framework for Temporal Forgery Localization},
    author={Zhang, Rui and Wang, Hongxia and Du, Mingshan and Liu, Hanqing and Zhou, Yang and Zeng, Qiang},
    booktitle={Proceedings of the 31st {ACM} International Conference on Multimedia},
    pages={8749--8759},
    year={2023},
    publisher={{ACM}},
}

@article{cai2023glitch,
  title = {Glitch in the Matrix: A Large Scale Benchmark for Content Driven Audio-Visual Forgery Detection and Localization},
  author = {Cai, Zhixi and Ghosh, Shreya and Dhall, Abhinav and Gedeon, Tom and Stefanov, Kalin and Hayat, Munawar},
  journal = {Computer Vision and Image Understanding},
  year = {2023},
  volume = {236},
  pages = {103818},
}

@article{kobyzev2020normalizing,
  title={Normalizing flows: An introduction and review of current methods},
  author={Kobyzev, Ivan and Prince, Simon JD and Brubaker, Marcus A},
  journal={IEEE transactions on pattern analysis and machine intelligence},
  volume={43},
  number={11},
  pages={3964--3979},
  year={2020},
  publisher={IEEE}
}

@misc{kingma2013auto,
  title={Auto-encoding variational bayes},
  author={Kingma, Diederik P and Welling, Max and others},
  year={2013},
  publisher={Banff, Canada}
}

@inproceedings{liu2015deep,
  title={Deep learning face attributes in the wild},
  author={Liu, Ziwei and Luo, Ping and Wang, Xiaogang and Tang, Xiaoou},
  booktitle={Proceedings of the IEEE international conference on computer vision},
  pages={3730--3738},
  year={2015}
}

@inproceedings{zhu2022celebv,
  title={CelebV-HQ: A large-scale video facial attributes dataset},
  author={Zhu, Hao and Wu, Wayne and Zhu, Wentao and Jiang, Liming and Tang, Siwei and Zhang, Li and Liu, Ziwei and Loy, Chen Change},
  booktitle={European conference on computer vision},
  pages={650--667},
  year={2022},
  organization={Springer}
}

@inproceedings{xie2022vfhq,
  title={Vfhq: A high-quality dataset and benchmark for video face super-resolution},
  author={Xie, Liangbin and Wang, Xintao and Zhang, Honglun and Dong, Chao and Shan, Ying},
  booktitle={Proceedings of the IEEE/CVF Conference on Computer Vision and Pattern Recognition},
  pages={657--666},
  year={2022}
}

@article{chung2018voxceleb2,
  title={Voxceleb2: Deep speaker recognition},
  author={Chung, Joon Son and Nagrani, Arsha and Zisserman, Andrew},
  journal={arXiv preprint arXiv:1806.05622},
  year={2018}
}

@article{kuznetsova2020open,
  title={The open images dataset v4: Unified image classification, object detection, and visual relationship detection at scale},
  author={Kuznetsova, Alina and Rom, Hassan and Alldrin, Neil and Uijlings, Jasper and Krasin, Ivan and Pont-Tuset, Jordi and Kamali, Shahab and Popov, Stefan and Malloci, Matteo and Kolesnikov, Alexander and others},
  journal={International journal of computer vision},
  volume={128},
  number={7},
  pages={1956--1981},
  year={2020},
  publisher={Springer}
}

@inproceedings{cheng2025taming,
  title={{MMAudio}: Taming Multimodal Joint Training for High-Quality Video-to-Audio Synthesis},
  author={Cheng, Ho Kei and Ishii, Masato and Hayakawa, Akio and Shibuya, Takashi and Schwing, Alexander and Mitsufuji, Yuki},
  booktitle={Proceedings of the Computer Vision and Pattern Recognition Conference},
  year={2025}
}

@article{tian2025audiox,
  title={AudioX: Diffusion Transformer for Anything-to-Audio Generation},
  author={Tian, Zeyue and Jin, Yizhu and Liu, Zhaoyang and Yuan, Ruibin and Tan, Xu and Chen, Qifeng and Xue, Wei and Guo, Yike},
  journal={arXiv preprint arXiv:2503.10522},
  year={2025}
}

@misc{kling2025,
author={KlingTeam},
 title="\url{https://klingai.com/global/}",
 year={2025}
}

@misc{pika2025,
author={PikaTeam},
 title="\url{https://pika.art/}",
 year={2025}
}

@misc{jimeng2025,
author={JimengTeam},
 title="\url{https://jimeng.jianying.com/ai-tool/home/}",
 year={2025}
}

@misc{ai2025magi1autoregressivevideogeneration,
      title={MAGI-1: Autoregressive Video Generation at Scale},
      author={Sand.ai and Hansi Teng and Hongyu Jia and Lei Sun and Lingzhi Li and Maolin Li and Mingqiu Tang and Shuai Han and Tianning Zhang and W. Q. Zhang and Weifeng Luo and Xiaoyang Kang and Yuchen Sun and Yue Cao and Yunpeng Huang and Yutong Lin and Yuxin Fang and Zewei Tao and Zheng Zhang and Zhongshu Wang and Zixun Liu and Dai Shi and Guoli Su and Hanwen Sun and Hong Pan and Jie Wang and Jiexin Sheng and Min Cui and Min Hu and Ming Yan and Shucheng Yin and Siran Zhang and Tingting Liu and Xianping Yin and Xiaoyu Yang and Xin Song and Xuan Hu and Yankai Zhang and Yuqiao Li},
      year={2025},
      eprint={2505.13211},
      archivePrefix={arXiv},
      primaryClass={cs.CV},
      url={https://arxiv.org/abs/2505.13211},
}

@inproceedings{miao2025mffi,
  title={MFFI: Multi-dimensional face forgery image dataset for real-world scenarios},
  author={Miao, Changtao and Zhang, Yi and Luo, Man and Feng, Weiwei and Zheng, Kaiyuan and Chu, Qi and Gong, Tao and Li, Jianshu and Diao, Yunfeng and Zhou, Wei and others},
  booktitle={Proceedings of the 33rd ACM International Conference on Multimedia},
  pages={13235--13242},
  year={2025}
}

@inproceedings{xi2023ai,
  title={Ai-generated image detection using a cross-attention enhanced dual-stream network},
  author={Xi, Ziyi and Huang, Wenmin and Wei, Kangkang and Luo, Weiqi and Zheng, Peijia},
  booktitle={2023 Asia Pacific Signal and Information Processing Association Annual Summit and Conference (APSIPA ASC)},
  pages={1463--1470},
  year={2023},
  organization={IEEE}
}

@article{ma2023iml,
  title={IML-ViT: Benchmarking image manipulation localization by vision transformer},
  author={Ma, Xiaochen and Du, Bo and Jiang, Zhuohang and Hammadi, Ahmed Y Al and Zhou, Jizhe},
  journal={arXiv preprint arXiv:2307.14863},
  year={2023}
}

@inproceedings{liu2021spatial,
  title={Spatial-phase shallow learning: rethinking face forgery detection in frequency domain},
  author={Liu, Honggu and Li, Xiaodan and Zhou, Wenbo and Chen, Yuefeng and He, Yuan and Xue, Hui and Zhang, Weiming and Yu, Nenghai},
  booktitle={Proceedings of the IEEE/CVF conference on computer vision and pattern recognition},
  pages={772--781},
  year={2021}
}

@inproceedings{ojha2023towards,
  title={Towards universal fake image detectors that generalize across generative models},
  author={Ojha, Utkarsh and Li, Yuheng and Lee, Yong Jae},
  booktitle={Proceedings of the IEEE/CVF conference on computer vision and pattern recognition},
  pages={24480--24489},
  year={2023}
}

@article{miao2025multi,
  title={Multi-spectral class center network for face manipulation localization},
  author={Miao, Changtao and Chu, Qi and Tan, Zhentao and Jin, Zhenchao and Gong, Tao and Zhuang, Wanyi and Wu, Yue and Liu, Bin and Hu, Honggang and Yu, Nenghai},
  journal={ACM Transactions on Multimedia Computing, Communications and Applications},
  volume={21},
  number={8},
  pages={1--24},
  year={2025},
  publisher={ACM New York, NY}
}

@inproceedings{miao2025mixture,
  title={Mixture-of-noises enhanced forgery-aware predictor for multi-face manipulation detection and localization},
  author={Miao, Changtao and Chu, Qi and Gong, Tao and Tan, Zhentao and Jin, Zhenchao and Zhuang, Wanyi and Luo, Man and Hu, Honggang and Yu, Nenghai},
  booktitle={Proceedings of the 33rd ACM International Conference on Multimedia},
  pages={3903--3912},
  year={2025}
}

@inproceedings{chen2026thinking,
  title={Thinking Like a Forensic Expert: A Multimodal Reasoning Chain for Training-Free Image Manipulation Localization},
  author={Chen, Rui and Liu, Bin and Miao, Changtao and Wang, Xinghao and Li, Yi and Gong, Tao and Chu, Qi and Yu, Nenghai},
  booktitle={2026 IEEE International Conference on Image Processing (ICIP)},
  pages={1--6},
  year={2026},
  organization={IEEE}
}

@article{wang2026sapl,
  title={SAPL: Semantic-Agnostic Prompt Learning in CLIP for Weakly Supervised Image Manipulation Localization},
  author={Wang, Xinghao and Miao, Changtao and Sheng, Dianmo and Gong, Tao and Chu, Qi and Yu, Nenghai and Zou, Quanchen and Zhang, Deyue and Zhang, Xiangzheng},
  journal={arXiv preprint arXiv:2601.06222},
  year={2026}
}

@article{che2026gifguard,
  title={GIFGuard: Proactive Forensics against Deepfakes in Facial GIFs via Spatiotemporal Watermarking},
  author={Che, Shupeng and Guo, Zhiqing and Miao, Changtao and Ma, Dan and Yang, Gaobo},
  journal={arXiv preprint arXiv:2604.26519},
  year={2026}
}

@article{li2026prototype,
  title={Prototype Memory-based Neighboring Feature Fusion Network for Image Manipulation Localization},
  author={Li, Songlin and Guo, Zhiqing and Miao, Changtao and Yang, Wenzhong and Wang, Liejun and Yang, Gaobo and Liao, Xin},
  journal={IEEE Transactions on Circuits and Systems for Video Technology},
  year={2026},
  publisher={IEEE}
}

@article{miao2023f,
  title={{F$^2$Trans}: High-Frequency Fine-Grained Transformer for Face Forgery Detection},
  author={Miao, Changtao and Tan, Zichang and Chu, Qi and Liu, Huan and Hu, Honggang and Yu, Nenghai},
  journal={IEEE Transactions on Information Forensics and Security},
  volume={18},
  pages={1039--1051},
  year={2023},
  publisher={IEEE}
}

@article{miao2022hierarchical,
  title={Hierarchical frequency-assisted interactive networks for face manipulation detection},
  author={Miao, Changtao and Tan, Zichang and Chu, Qi and Yu, Nenghai and Guo, Guodong},
  journal={IEEE Transactions on Information Forensics and Security},
  volume={17},
  pages={3008--3021},
  year={2022},
  publisher={IEEE}
}

@article{miao2021learning,
  title={Learning forgery region-aware and ID-independent features for face manipulation detection},
  author={Miao, Changtao and Chu, Qi and Li, Weihai and Li, Suichan and Tan, Zhentao and Zhuang, Wanyi and Yu, Nenghai},
  journal={IEEE Transactions on Biometrics, Behavior, and Identity Science},
  volume={4},
  number={1},
  pages={71--84},
  year={2022},
  publisher={IEEE}
}

@article{diao2026vulnerabilities,
  title={Vulnerabilities in ai-generated image detection: The challenge of adversarial attacks},
  author={Diao, Yunfeng and Zhai, Naixin and Miao, Changtao and Yu, Zitong and Wei, Xingxing and Yang, Xun and Wang, Meng},
  journal={IEEE Transactions on Multimedia},
  year={2026},
  publisher={IEEE}
}











\end{document}